\documentclass[11pt,a4paper]{article}
\usepackage[hyperref]{emnlp2020}
\usepackage{times}
\usepackage{latexsym}
\usepackage{subfig}
\usepackage{graphicx}
\usepackage[symbol]{footmisc}

\usepackage{microtype}

\aclfinalcopy 

\title{Cross-lingual Alignment Methods for Multilingual BERT:\\ A Comparative Study}

\author{Saurabh Kulshreshtha\thanks{\hspace{0.2cm}Work done during an internship at Amazon.}\\
  Department of Computer Science\\
  University of Massachusetts Lowell \\
  \texttt{skul@cs.uml.edu} \\\And
  José Luis Redondo-García \\
  Amazon Alexa \\
  Cambridge, UK \\
  \texttt{jluisred@amazon.com} \\\And
  Ching-Yun Chang \\
  Amazon Alexa \\
  Cambridge, UK \\
  \texttt{cychang@amazon.com} \\}

\date{}

\begin{document}
\maketitle

\begin{abstract}
Multilingual BERT (mBERT) has shown reasonable capability for zero-shot cross-lingual transfer when fine-tuned on downstream tasks. Since mBERT is not pre-trained with explicit cross-lingual supervision, transfer performance can further be improved by aligning mBERT with cross-lingual signal. Prior work proposes several approaches to align contextualised embeddings. In this paper we analyse how different forms of cross-lingual supervision and various alignment methods influence the transfer capability of mBERT in zero-shot setting. Specifically, we compare parallel corpora vs.\ dictionary-based supervision and rotational vs.\ fine-tuning based alignment methods. We evaluate the performance of different alignment methodologies across eight languages on two tasks: Name Entity Recognition and Semantic Slot Filling. In addition, we propose a novel normalisation method which consistently improves the performance of rotation-based alignment including a notable 3\% F1 improvement for distant and typologically dissimilar languages. Importantly we identify the biases of the alignment methods to the type of task and proximity to the transfer language. We also find that supervision from parallel corpus is generally superior to dictionary alignments.
\end{abstract}

\section{Introduction}
Multilingual BERT (mBERT) ~\cite{devlin} is the BERT architecture trained on data from 104 languages where all languages are embedded in the same vector space. Due to the multilingual and contextual representation properties of mBERT, it has gained popularity in various multilingual and cross-lingual tasks ~\cite{Karthikeyan2020CrossLingualAO, wu}. In particular, it has demonstrated good zero-shot cross-lingual transfer performance on many downstream tasks, such as Document Classification, NLI, NER, POS tagging, and Dependency Parsing~\cite{wu}, when the source and the target languages are similar.

Many experiments \cite{ahmad-etal-2019-difficulties} suggest that to achieve reasonable performance in the zero-shot setup, the source and the target languages need to share similar grammatical structure or lie in the same language family. In addition, since mBERT is not trained with explicit language signal, mBERT's multilingual representations are less effective for languages with little lexical overlap ~\cite{patra}. One branch of work is therefore dedicated to improve the multilingual properties of mBERT by aligning the embeddings of different languages with cross-lingual supervision.

Broadly, two methods have been proposed in prior work to induce cross-lingual signals in contextual embeddings: 1) Rotation Alignment as described in Section \ref{rotation_based_alignment} aims at learning a linear rotation transformation to project source language embeddings into their respective locations in the target language space ~\cite{schuster2019,zwang,diab}; 2) Fine-tuning Alignment as explained in Section \ref{fine_tuning_based_alignment} internally aligns language sub-spaces in mBERT through tuning its weights such that distances between embeddings of word translations decrease while not losing the informativity of the embeddings ~\cite{cao}. Additionally, two sources of cross-lingual signal have been considered in literature to align languages: parallel corpora and bilingual dictionaries. While the choice of each alignment method and source of supervision have a variety of advantages and disadvantages, it is unclear how these affect the performance of the aligned spaces across languages and various tasks.

In this paper, we empirically investigate the effect of these cross-lingual alignment methodologies and applicable sources of cross-lingual supervision by evaluating their performance on zero-shot Named Entity Recognition (NER), a structured prediction task, and Semantic Slot-filling (SF), a semantic labelling task, across eight language pairs.

The motivation for choice of these tasks to evaluate are two-fold: 1. Prior work has already studied alignment methods on sentence level tasks. \newcite{cao} show the effectiveness of mBERT alignment methods on XNLI \shortcite{xnli}. 2. Word-level tasks do not benefit from more pre-training unlike other language tasks that improve by simply supplementing with more pre-training data. In experiments over the XTREME benchmark, \newcite{xtreme} find that transfer performance improves across all tasks when multilingual language models are pre-trained with more data, with the sole exception of word-level tasks. They note that this indicates current deep pre-trained models do not fully exploit the pre-training data to transfer to word-level tasks. We believe that NER and Slot-filling tasks are strong candidate tasks to assess alignment methods due to limited cross-lingual transfer capacity of current models to these tasks.

To the authors' knowledge, this is the first paper exploring the comparison of alignment methods for contextual embedding spaces: rotation vs.\ fine-tuning alignment and two sources of cross-lingual supervision: dictionary vs. parallel corpus supervision on a set of tasks of structural and semantic nature over a wide range of languages. From the results, we find that parallel corpora are better suited for aligning contextual embeddings. In addition, we find that rotation alignment is more robust for primarily structural NER downstream tasks while the fine-tuning alignment is found to improve performance across semantic SF tasks. In addition, we propose a novel normalisation procedure which consistently improves rotation alignment, motivated by the structure of mBERT space and how languages are distributed across it. We also find the effect of language proximity on transfer improvement for these alignment methods.

\section{Rotation-based Alignment}\label{rotation_based_alignment}
 \newcite{mikolov} proposed to learn a linear transformation $W_{s\,\to\,t}$ which would project an embedding in the source language $e_s$ to its translation in the target language space $e_t$, by minimising the distances between the projected source embeddings and their corresponding target embeddings:
\begin{equation}
\label{eqn:mikolov}
\min_{W \in {R}^{d \times d}} \left\|W X_{s}-X_{t} \right\|
\end{equation}
$X_{s}$ and $X_{t}$ are matrices of size $d \times K$ where $d$ is the dimensionality of embeddings and $K$ is the number of parallel words from word-aligned corpora, or word pairs from a bilingual dictionary between the source and target languages. Further work~\newcite{xing} demonstrated that restricting $W$ to a purely rotational transform improves cross-lingual transfer across similar languages. The orthogonality assumption reduces Eq.(\ref{eqn:mikolov}) into the so-called Procrustes problem with the closed form solution:
\begin{equation}
\label{eqn:procrustus1}
W=U V^{T},
\end{equation}
\begin{equation}
\label{eqn:procrustus2}
\textrm{ where } U \Sigma V^{T}=SVD\left(X_{t} X_{s}^{T}\right)
\end{equation}
and the SVD operator stands for Singular Value Decomposition.

\subsection{Language Centering Normalization}\label{language_centering}

A purely rotational transformation can align two embedding spaces only if the two spaces are roughly isometric and are distributed about the same mean. In case the two embedding distributions are not centered around the same mean, meaning the two spaces have little overlap and are shifted by a translation offset in the space, they cannot be aligned solely through rotation.

Since the linear transformation $W_{s\,\to\,t}$ derived from solving the Procrustus problem only rotates the vector space, it assumes the embeddings of two languages are zero-centered. However \newcite{libovick2019languageneutral} observe that languages distributions in mBERT have distinct and separable centroids and different language families have well separated sub-spaces in the mBERT embedding vector space. To address this discrepancy, we propose a new normalisation mechanism which entails:

Step 1. Normalising the embeddings of both languages so that they have zero mean:
\begin{equation}
\label{eqn:norm1}
\hat{X}_s = X_s - \bar{X}_s \textrm{ and } \hat{X}_t = X_t - \bar{X}_t
\end{equation}
where $\bar{X}_s$ and $\bar{X}_t$ are centroids of source and target embeddings $X_s$ and $X_t$; and $\hat{X}_s$ and $\hat{X}_t$ are mean-centered source and target language embeddings their rows correspond to word translations. Next, $\hat{X}_s$ and $\hat{X}_t$ are used to compute the transformation matrix $\hat{W}_{s\,\to\,t}$ by solving Eq.(\ref{eqn:procrustus1}) and Eq.(\ref{eqn:procrustus2}).

Step 2. During training a downstream task, embedding of a source language word $e_s$ needs to be re-centered, rotated and finally translated to the target language subspace to derive the projection $e_{t^*}$:
\begin{equation}
\label{eqn:norm2}
e_{t^*} = \hat{W}_{s\,\to\,t} (e_s - \bar{X}_s) + \bar{X}_t
\end{equation}

This helps the task specific model, particularly in zero-shot setting, by projecting the source language task data to the same locality as the target language.

\begin{figure*}[ht]
\begin{tabular}{cc}
\includegraphics[width=.48\linewidth]{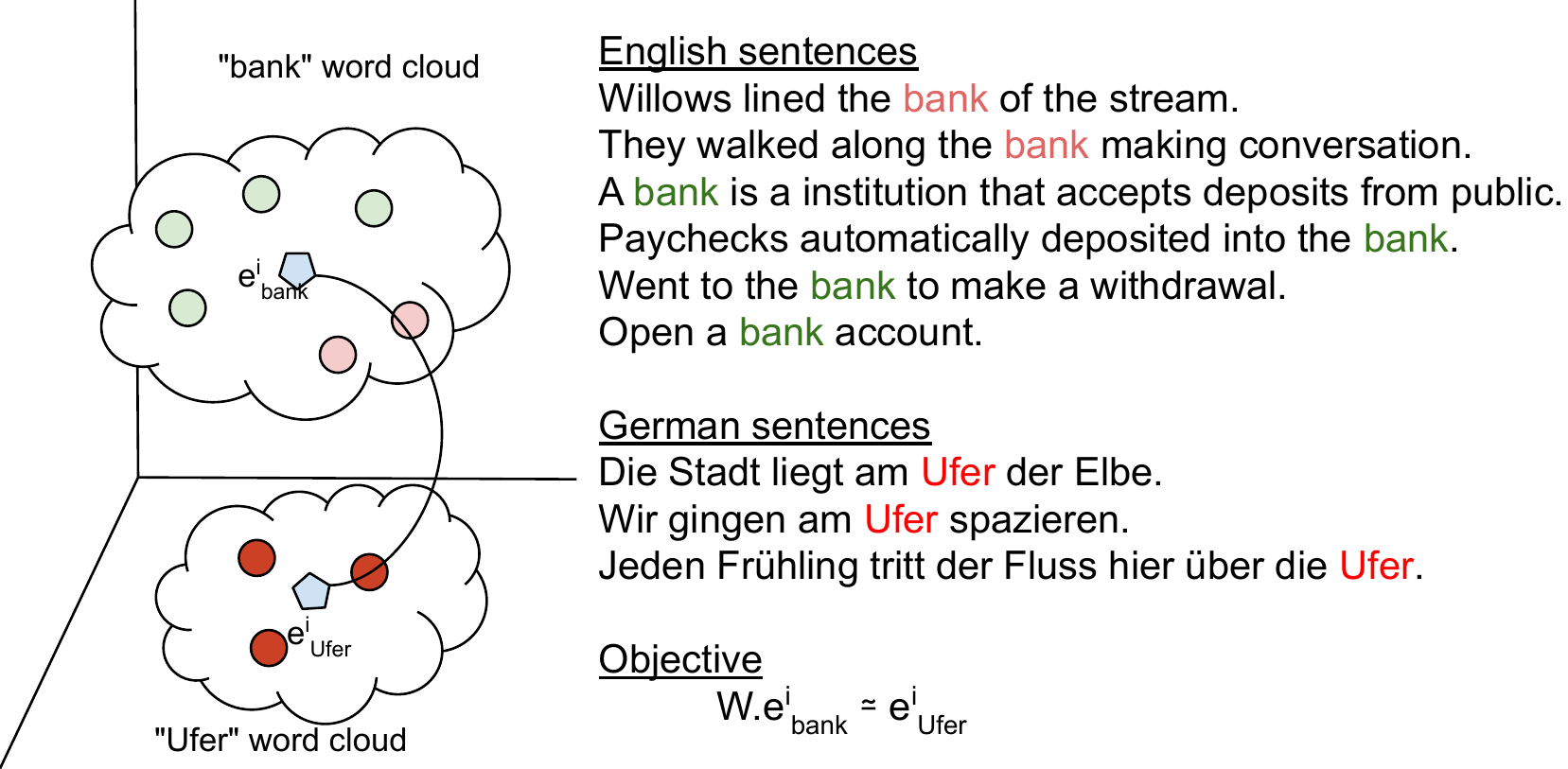} & \includegraphics[width=.48\linewidth]{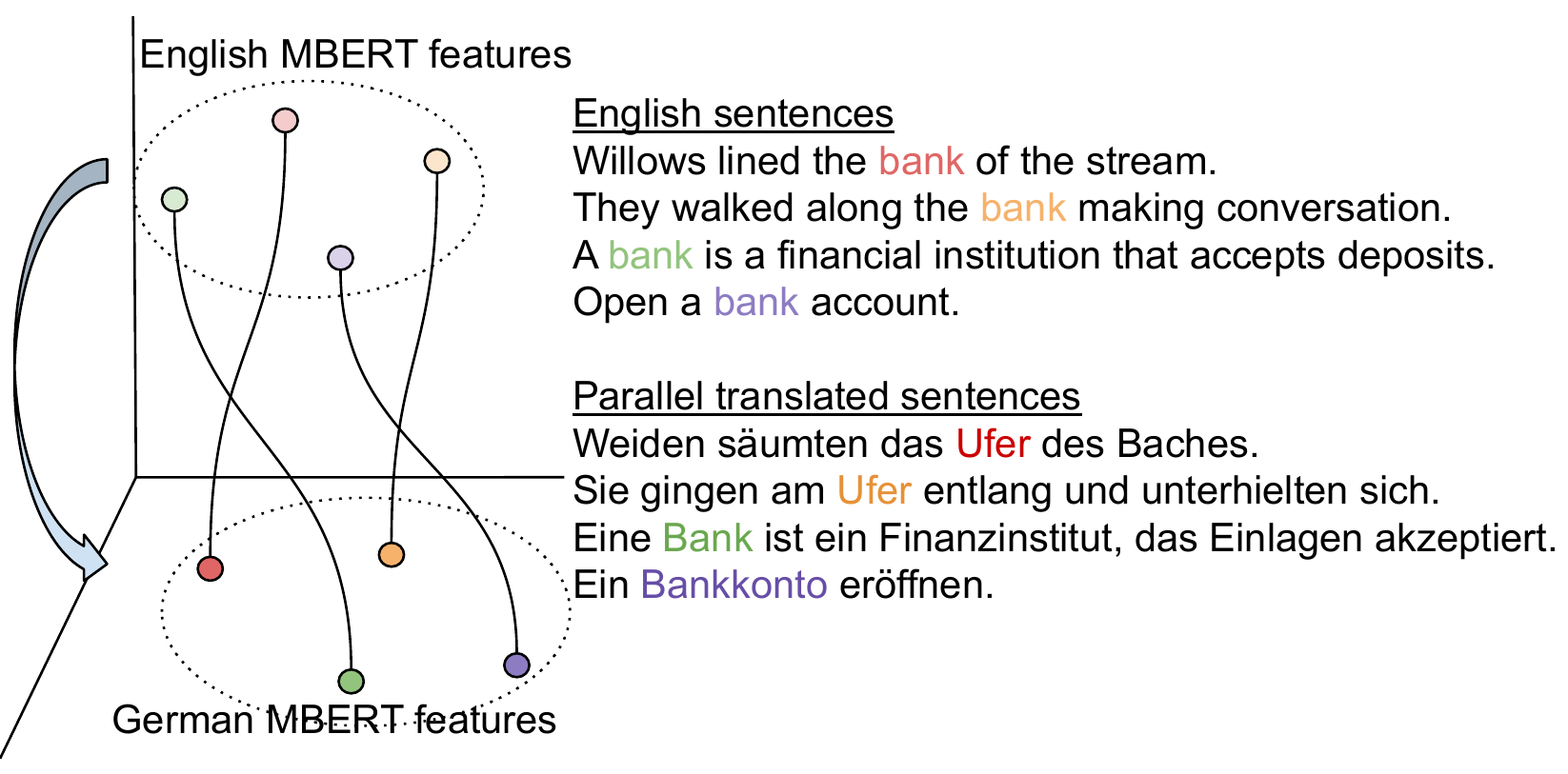} \\
(a) Alignment with dictionary. & (b) Alignment with parallel sentences.
\end{tabular}
\caption{\label{fig:supervision} In Figure 1a, contextual embeddings of the word ``bank'' get averaged across all word senses noted by different colors into single word anchor embedding. Figure 1b illustrates supervision from parallel corpora where word-alignments correspond to translation in similar context noted by similar colors (lighter for English), this provides more fine-grained supervision for contextualised alignment of mBERT. }
\end{figure*}
\subsection{Supervision Signals for Rotation Alignment}
In this section we describe how existing work utilises two different cross-lingual signals, bilingual dictionaries and parallel corpora, to supervise rotation alignment. Additionally, we analyse the advantages and disadvantages of the two choices.

\subsubsection{Bilingual Dictionary Supervision}\label{sec:bilingual_dictionary}
In order to utilise a bilingual dictionary to supervise the embedding alignment, each word in the dictionary needs to have a single representation. However the same word can have many representations in the contextualised language model vector space depending on the context it occurs in. \newcite{schuster2019} observes that the contextual embeddings of the same word form a tight cluster - word cloud, the centroid of this word cloud is distinct and separable for individual words. They further propose that centroid of a word cloud can be considered as the context-independent representation of a word, called average word anchor. These word anchors are computed by averaging embeddings over all occurrences of a word in a monolingual corpora, where words occur in a variety of contexts. Formally the mBERT embedding of a source language word $s_m$ in context $c_h$ is denoted as $e_{s_m, c_h}$. If this word occurs a total of $p$ times in the monolingual corpus, that is in contexts ${c_1, c_2, ... c_p}$, the anchor word embedding $A_{s_m}$ for word $s_m$ across all the contexts is the average:

\begin{equation}
\label{eqn:schuster}
A_{s_m} = \frac{\displaystyle\sum_{h=1}^{p} e_{s_m, c_h}}{p}
\end{equation}

Average word anchor pair $(A^i_{s_m}, A^i_{t_{m^*}})$ , where $i$ is the mBERT layer, for all word pairs from the dictionary $(s_m, t_{m^*})$  form the rows of matrices $X^i_s$ and $X^i_t$ respectively, which are then used to solve Eq.(\ref{eqn:procrustus1}) and Eq.(\ref{eqn:procrustus2}), resulting in an alignment transformation matrix $W^i_{s\,\to\,t}$.

However, there are limitations to this approach. \newcite{multisense} found that the word cloud of multi-sense words, such as the word ``bank'', which can mean either the financial institution or the edge of a river depending on the context, are further composed of clearly separable clusters, for every word sense. Averaging over multiple contextual embeddings infers losing certain degree of contextual information at both the source and target language words. Figure~\ref{fig:supervision}a visualises word anchor calculation and also highlights this limitation. On the other hand, one of the advantages of this method is that bilingual dictionaries are available for even very low resource languages.

\subsubsection{Parallel Corpus Supervision}
\label{sec:rot_parallel}
Word-aligned parallel sentences can be utilised as a source of cross-lingual signal to align contextual embeddings ~\cite{diab,zwang}. Given a parallel corpora, $s_m$ and $t_{m^*}$ are aligned source and the target language words appearing in context $c_h$ and $c_{h^*}$, respectively. The parallel word embedding matrices $X^i_s$ and $X^i_t$ for mBERT layer-$i$ are composed from the contextual embeddings $e^i_{s_m, c_h}$ and $e^i_{t_{m^*}, c_{h^*}}$ respectively, and are used to solve Eq.(\ref{eqn:procrustus1}) and Eq.(\ref{eqn:procrustus2}) to derive an alignment transformation matrix $W^i_{s\,\to\,t}$. 

Figure~\ref{fig:supervision}a and \ref{fig:supervision}b illustrate how parallel supervision is more suited to align contextual embeddings compared to dictionary supervision where multiple senses of a word are compressed into a single word anchor. However, parallel corpora rarely come with word-alignment annotations that are often automatically generated by off-the-shelf tools such as \textit{fast\_align}~\cite{fastalign}, which can be noisy. It is worth noting that word alignment error rate of an off-the-shelf tool drops when number of parallel sentences increases, therefore parallel corpus supervision is favourable for languages where more parallel data is available.

\section{Fine-tuning Alignment with Parallel Corpora}\label{fine_tuning_based_alignment}
Rotation alignment has a strong assumption that the two language spaces (or sub-spaces in case of mBERT) are approximately isometric~\cite{sogaard2018limitations}. \newcite{patra} reported that the geometry of language embeddings becomes dissimilar for distant languages, and the isometry assumption degrades the alignment performance in such cases. In addition, as explained in Section~\ref{language_centering} rotation alignment alone cannot achieve effective mapping when two languages spaces have separate centroids. Therefore, next we consider existing work to non-linearly align two language spaces.

\newcite{cao} proposed to directly align languages within mBERT model through fine-tuning. The objective of the fine-tuning is to minimise the distance between the two contextual representations of an aligned word pair in parallel corpora:
\begin{equation}
\label{eqn:finetune-align}
L^i_{align} = min \displaystyle\sum_{m,m^*}\| e^i_{s_m} - e^i_{t_{m^*}}\|
\end{equation}
However, fine-tuning with only the above objective would led to lose the semantic information in mBERT learnt during pre-training, since a trivial solution to the Eq.(\ref{eqn:finetune-align}) can be simply to make all the embeddings equal. To deal with this, \newcite{cao} also proposed a regularisation loss that does not allow the embedding of a source language word to stray too far away from its original location $\textbf{e}^{i}_{s_m}$ in the pre-trained mBERT model, namely:
\begin{equation}
\label{eqn:finetune-regularize-1}
L^i_{regularise} = min \displaystyle\sum_{m}\| e^{i}_{s_m} - \textbf{e}^{i}_{s_m} \|
\end{equation}
Note that $\textbf{e}^{i}_{s_m}$ is generated from a copy of the original pre-trained mBERT model where the parameters are kept frozen. Both of the alignment and the regularization losses are combined and jointly optimised in order to align the two language subspaces while maintaining informativity of embeddings:
\begin{equation}
\label{eqn:finetune}
L_{finetune} = min \displaystyle\sum_{i=n_s}^{n_e} L^i_{align} + L^i_{regularise}
\end{equation}
Here $n_s$ to $n_e$ is the range of mBERT layers aligned. We experimented with two variants of the fine-tuning approach: 1) moving target language towards source language while keeping the source embeddings approximately fixed through the regularization term in Eq.(\ref{eqn:finetune-regularize-1}); 2) moving the source language embeddings towards the target space while keeping the target language space relatively fixed, then the regularisation loss changes to:
\begin{equation}
\label{eqn:finetune-regularize-2}
L^i_{regularise} = min \displaystyle\sum_{m^*}\| e^{i}_{t_{m^*}} - \textbf{e}^{i}_{t_{m^*}} \|
\end{equation}

\section{Experimental Setup}
In this section, we firstly describe the resources and implementation details of the alignment methods followed by the zero-shot NER and SF tasks used to evaluate the alignments. In addition, we briefly explain the datasets used in the experiments.
\subsection{Learning Alignments}
Our baseline model is a pre-trained mBERT\footnote{Available for download at:  \url{https://github.com/google-research/bert/blob/master/multilingual.md}} -- 12 transformer layers, 12 attention heads, 768 hidden dimensions -- denoted as \emph{mBERT Baseline}. When a word is tokenised into multiple subwords by the tokeniser, we average their corresponding subword embeddings to obtain embedding for the word. Following \newcite{zwang} we collect 30k parallel sentences for each of the language pairs from publicly available parallel corpora. For the European languages, German, Italian, Spanish and Dutch, the Europarl corpus~\cite{europarl} is used; for Hindi, Turkish and Thai, the OpenSubtitles corpus~\cite{opensubtitles} is used; for Armenian the parallel sentences are extracted from the QED Corpus~\cite{qed}. We obtain contextual and  average anchor embeddings described in Section~\ref{sec:bilingual_dictionary} by passing the corpora described above through pre-trained mBERT.

We use the bilingual dictionaries provided with the MUSE framework ~\cite{muse} as the source for dictionary supervision. As for the parallel corpus supervision, since none of the collected parallel sentences contains word-level alignment information, we utilise \textit{fast\_align}~\cite{fastalign} to automatically derive word alignment signals.

For the rotation alignment, we compute four independent transformation matrices for each of the last four transformer layers similar to \newcite{zwang}. We use \emph{RotateAlign} and \emph{NormRotateAlign} to refer the rotation alignment learnt without and with the proposed language centering normalisation, respectively. To be consistent, for the fine-tuning alignment we align the word representations in the last four transformer layers of the mBERT model, denoted as \emph{FineTuneAlign}.

\begin{table*}
\begin{center}
\resizebox{\textwidth}{!}{%
\begin{tabular}{|l|l|l|l|l|l|}
\hline
Datasets & Task & Translated & \# Language \& Train/Dev/Test Size & \# Slot Types & Domains \\ \hline
\begin{tabular}{l}
CoNLL\shortcite{conll2002,conll2003} \\ PioNER${}^1$\shortcite{pioner} \\
\end{tabular} & NER & No & \
\begin{tabular}{ll}
\textbf{en} & 14,987 / 3,466 / 3,684 \\
de & 12,705 / 3,068 / 3,160 \\
es & 8,323 / 1,915 / 1,517 \\
nl & 15,806 / 2,895 / 5,195 \\
hy & 5,964 / 1,491 / 2,529
\end{tabular}
& 4 & News Articles \\ \hline
\begin{tabular}{l}
ATIS\shortcite{atis}\\
ATIS-HI,TK\shortcite{atis-google}
\end{tabular}
 & SF & Yes & \begin{tabular}{ll}
\textbf{en} & 4,478 / 500 / 893 \\
hi & 600 / 893 / 893 \\
tk & 600 / 715 / 715
\end{tabular} & 63 & Air Travel \\ \hline
\begin{tabular}{l}
FB\shortcite{fb}
\end{tabular}
 & SF & Yes & \begin{tabular}{ll}
\textbf{en} & 30,521 / 4,181 / 8,621 \\
es & 3,617 / 1,983 / 3,043 \\
th & 2,156 / 1,235 / 1,692
\end{tabular} & 11 & Weather, Alarm, Reminder \\ \hline
\begin{tabular}{l}
SNIPS\shortcite{snips} \\ Almaware-SLU\shortcite{snips-it} \\
\end{tabular}& SF & Yes & \begin{tabular}{ll}
\textbf{en} & 13,084 / 700 / 700 \\
it & 1,400 / 700 / 700
\end{tabular} & 39 & \begin{tabular}[c]{@{}l@{}}Music, Restaurants, TV, Movies,\\ Books, Weather\end{tabular} \\ \hline
\end{tabular}}
\end{center}
\caption{\label{table:datasets} Summary of NER and SF dataset families. English marked in bold is treated as the source language.}
\end{table*}

\begin{table*}[]
\centering
\small
\begin{tabular}{|c|c|}
\hline
CoNLL-NER & [U.N.]${}_{ORG}$ official [Ekeus]${}_{PER}$ heads for
[Baghdad]${}_{LOC}$.\\\hline
ATIS-SF & show the [latest]${}_{flight\_mod}$ flight from [denver]${}_{fromloc.city\_name}$ to [boston]${}_{toloc.city\_name}$ \\\hline
FB-SF & do you have [wednesday's]${}_{datetime}$ [weather forecast]${}_{weather\_noun}$ for [half moon bay]${}_{location}$ \\\hline
SNIPS-SF & add this [track]${}_{music\_item}$ to [my]${}_{playlist\_owner}$ [global funk]${}_{playlist}$ \\\hline
\end{tabular}
\caption{\label{table:examples} Examples from the datasets.}
\end{table*}

\subsection{Evaluation of the Alignments}

We evaluate the learnt alignments using two downstream tasks: Named Entity Recognition (NER) and Semantic Slot Filling (SF), both of which aim to predict a label for each token in a sentence. NER is a more structural task with fewer entity types and involves less semantic understanding of the context compared to SF. Examples of the tasks can be found in Table~\ref{table:examples}.

We use the same model architecture and hyper-parameters as \newcite{zwang}, two BiLSTM layers followed by a CRF layer, where learning rate is set to $10^{-4}$ for European languages and $10^{-5}$ for the other languages determined by the validation set. In order to measure the effectiveness of a learnt alignment, all the experiments are conducted with zero-shot settings similar to \newcite{zwang}, where the source language data is first transformed to the target language space and then used to train a BiLSTM-CRF model. The target language validation set is used for hyper-parameter tuning and reporting the evaluation results. For each experiment we report F1 scores averaged over 5 runs.

\subsection{NER and SF Datasets}

We use the following four families of datasets, each of which has the same set of labels. A summary of the datasets can be found in Table ~\ref{table:datasets}. Example utterances and annotations and shown in Table ~\ref{table:examples}.

\noindent \textbf{CoNLL-NER}: This includes CoNLL 2002, 2003 NER benchmark task~\cite{conll2002,conll2003} containing entity annotations for news articles in English, German, Spanish and Dutch. We also include in this family PioNER\footnote{PioNER data only has PER, LOC and ORG labels and does not contain MISC.}~\cite{pioner}, a manually annotated dataset in Armenian, which is typographically different from the other languages in this family. In this dataset-family, target language data is sourced from local news articles, and not generated through translation from source data.

\noindent \textbf{ATIS-SF}: ATIS Corpus~\cite{atis} is an English dataset containing conversational queries about flight booking. \newcite{atis-google} manually translated a subset of the data into two languages, Turkish and Hindi, along with crowd-sourced phrase-level annotations.

\noindent \textbf{FB-SF}: \newcite{fb} introduced Multilingual Task-Oriented Dialog Corpus in English, Spanish and Thai across three domains: weather, alarm and reminders, where Spanish and Thai data were manually translated and annotated from a subset of the English data.

\noindent \textbf{SNIPS-SF}: A multi-domain slot-filling dataset in English released by~\newcite{snips}. \newcite{snips-it} automatically translated this dataset into Italian, and then manually labelled the translation where entities were substituted by Italian entities collected from the Web.

\begin{table*}[t]
\begin{center}
\resizebox{\textwidth}{!}{%
\begin{tabular}{|l|c|c|c|c|c|c|c|c|c|}
\hline
\multicolumn{1}{|c|}{Dataset-Task} & \multicolumn{4}{c|}{CoNLL-NER} & \multicolumn{2}{c|}{ATIS-SF} & \multicolumn{2}{c|}{FB-SF} & SNIPS-SF \\ \hline
\multicolumn{1}{|c|}{Transfer Pair} & en to de & en to nl & en to es & en to hy & en to hi & en to tk & en to es & en to th & en to it \\ \hline
\multicolumn{10}{|c|}{Baselines from Literature} \\ \hline
mBERT \cite{wu} & 69.56 & 77.75 & 74.96 & - & - & - & - & - & - \\ \hline
mBERT Rotation Alignment: Parallel~\cite{zwang} & 70.54 & 79.03 & 75.77 & - & - & - & - & - & - \\ \hline
BERT, 1400 Target Language Train ~\cite{snips-it}$^\dagger$  & - & - & - & - & - & - & - & - & 83.04 \\ \hline
Non-contextual Zero-shot Baseline~\cite{atis-google}${}^*$ & - & - & - & - & $\sim$40 & $\sim$40 & - & - & - \\ \hline
Translate train~\cite{fb}${}^\ddag$ & - & - & - & - & - & - & 72.87 & 55.43 & - \\ \hline
\multicolumn{10}{|c|}{Our Experiments} \\ \hline
mBERT Baseline & 66.15 & 77.55 & 74.80 & 62.38 & 50.84 & 21.15 & 74.66 & 9.58 & 76.70 \\ \hline
RotateAlign${}_{dict}$ & 67.20 & 78.07 & 75.08 & - & 57.32 & 31.46 & 73.28 & 9.23 & 76.51 \\ \hline
NormRotateAlign${}_{dict}$ & 68.56 & 78.53 & 75.22 & - & \textbf{57.86} & 33.62 & 74.52 & 12.38 & 76.82 \\ \hline
RotateAlign${}_{parallel}$ & 70.48 & 79.52 & 75.84 & 65.31 & 52.24 & 37.38 & 73.57 & 9.12 & 77.70 \\ \hline
NormRotateAlign${}_{parallel}$ & \textbf{71.23} & \textbf{79.90} & \textbf{75.93} & \textbf{66.56} & 53.03 & 38.18 & 74.73 & 11.88 & 77.87 \\ \hline
FineTuneAlign${}_{tgt\to src}$ & 70.25 & 77.10 & 73.92 & 63.53 & 51.35 & \textbf{45.98} & 73.44 & 13.45 & 77.74 \\ \hline
FineTuneAlign${}_{src\to tgt}$ & 66.91 & 77.21 & 74.49 & 62.29 & 50.51 & 39.43 & \textbf{80.90} & \textbf{20.77} & \textbf{80.21} \\ \hline
\end{tabular}}
\end{center}
\caption{\label{table:results} Performance (F1 score) of the alignment methods on the zero-shot NER and SF tasks. Top scores within our experiments are marked in bold. No results are reported for Armenian dictionary alignments since English-Armenian dictionary was available in the MUSE framework. ${}^\dagger$ \newcite{snips-it} use 1400 Italian instances as part of the training data. ${}^{*}$ Numbers read from a chart in the paper. ${}^\ddag$ \newcite{fb} uses a machine translation model to translate this dataset and word alignments generated by attention weights to infer annotation.}
\end{table*}

\section{Results and Analysis}

The evaluation results of each alignment method on the downstream NER and SF tasks are reported in Table \ref{table:results} and Figure~\ref{fig:results}. In addition to the \emph{mBERT Baseline} and for comparison purposes, we also list relevant results found in literature~\cite{wu,zwang,atis-google,fb,snips-it} that have been evaluated on the same datasets.

\subsection{mBERT Baseline and Language Proximity}
\emph{mBERT Baseline} numbers can be indicative of how well languages are already aligned in the mBERT space. High zero-shot scores for German, Dutch, Spanish and Italian indicate that European languages are extremely well aligned to English in mBERT. However, distant languages such as Thai and Turkish, which belong to different language families (Kra–Dai and Turkic) than English, have poor alignment with low F1 scores of 9.58 and 21.15, respectively. Finally, moderately distant languages such as Armenian and Hindi, which fall within the larger Indo-European language family, have moderate alignment with English with scores of 62.38 and 50.84, respectively.

\subsection{mBERT Baseline vs./ Rotation Alignment}
\emph{RotateAlign} improves performance by 19\% absolute for ATIS-Turkish, going from baseline of 21.15 to 38.18 in F1 score. For ATIS-Hindi the performance improves from 50.84 to 57.86 F1 (7 points), and 4\% absolute for the PioNER-Armenian from 62.38 to 66.56. These numbers show how \emph{RotateAlign} can improve performance over \emph{mBERT Baseline} for moderately-close languages such as Hindi, Turkish and Armenian, while there is only around 1 point improvement for European languages. This implies that Hindi, Turkish and Armenian subspaces are geometrically similar to English, however they are misaligned in terms of rotation in \emph{mBERT Baseline}.

However, in the case of Thai, which is a distant language from English, \emph{RotateAlign} does not improve performance over the \emph{mBERT Baseline}. This suggests that Thai and English's embedding spaces are structurally dissimilar.

\subsection{Rotation Alignment with vs./ without Language Centering Normalisation}
Applying the proposed language centering normalisation in Section~\ref{language_centering} before performing the rotation alignment, namely \emph{NormRotateAlign} in Table~\ref{table:results}, is found to further improve downstream performance across all tasks and languages. The improvement over \emph{RotateAlign} is up to 3\% absolute F1 for Thai, around 1\% absolute for moderately closer languages like Hindi, Turkish and Armenian, and around 0.5\% absolute F1 for closer target languages such as German. Note that Thai, which does not benefit from rotation alignment alone, improves by an average of 2.3 points after applying the normalisation. These results corroborate that language families that are further away from each other have more separable sub-spaces in the \emph{mBERT Baseline}, and bringing the language distributions closer helps the downstream task's performance. 

\subsection{Parallel Corpus vs./ Dictionary Supervision}
Amongst the cases where \emph{RotateAlign} improves performance over the \emph{mBERT Baseline}, parallel-corpus supervised \emph{RotateAlign} is superior to dictionary supervision, with the exception of Hindi. This could be explained by the fact that word anchors are independent of multiple word senses, thereby the cross-lingual signal is poorer compared to parallel word alignments. This is in line with observations from \newcite{multisense}.

\subsection{Rotation vs./ Fine-tuning Alignment}
\begin{figure}[ht]
  \includegraphics[width=0.47\textwidth]{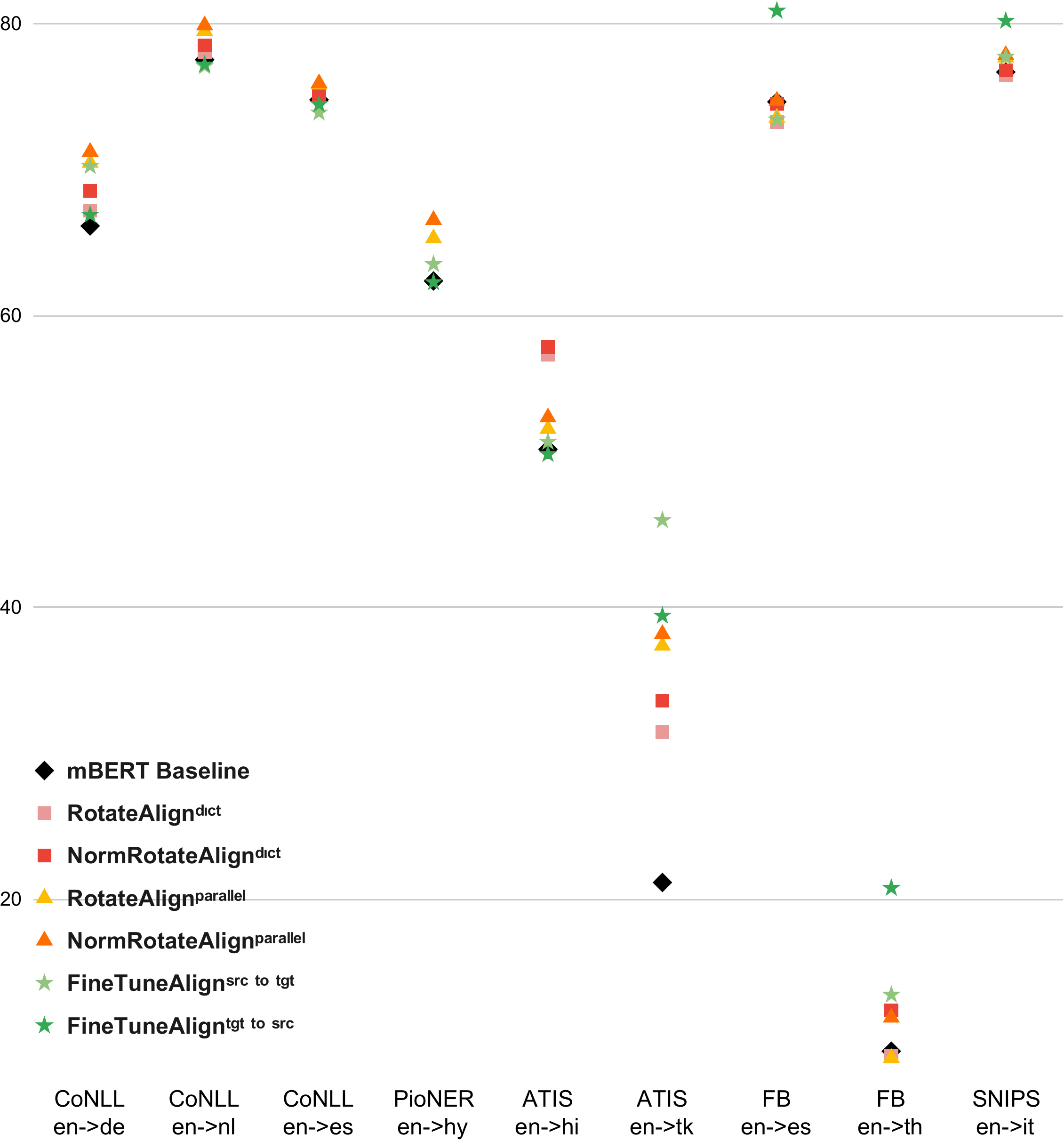}
  \caption{\label{fig:results}Trend of improvement from various alignment methods. Rotation alignment improves performance for NER, while fine-tuning alignment is found to be better for SF tasks. Improvements increase initially with distance between source and target languages and diminish for distant languages.}
\end{figure}

From Table~\ref{table:results} and Figure~\ref{fig:results} we can see that \emph{FineTuneAlign} explained in Section~\ref{fine_tuning_based_alignment} improves performance over \emph{RotateAlign} for semantic tasks (SF), with the only exception of ATIS-Hindi. On the other hand, \emph{FineTuneAlign} underperforms \emph{RotateAlign} for structural tasks (NER), and in some cases even fall behind \emph{mBERT Baseline}. Note that we notice no clear trend between \emph{FineTuneAlign${}_{src \to tgt}$} and \emph{FineTuneAlign${}_{tgt \to src}$}.

\emph{FineTuneAlign${}_{src \to tgt}$} improves over the best rotation alignment \emph{NormRotateAlign$_{parallel}$} by 7.8\% absolute for the ATIS-Turkish task from 38.18 to 45.98. It significantly outperforms \emph{mBERT Baseline} by 24 points. For FB-Thai \emph{FineTuneAlign${}_{src \to tgt}$} surpasses \emph{NormRotateAlign$_{dict}$} by 8.39\% absolute F1 from 12.38 to 20.77, 11 points higher than \emph{mBERT Baseline}. For FB-Spanish we observe an improvement from 74.73 to 80.90 (6\% absolute) compared to \emph{RotateAlign} and similarly +6 points compared to \emph{mBERT Baseline}. For SNIPS-Italian, \emph{FineTuneAlign} improves performance over \emph{NormRotateAlign} from 77.87 to 80.21 (2.5 points) and is 3.5 points better than \emph{mBERT Baseline}.

All SF tasks considered are generated by translation from the source language data. This may indicate that the fine-tuning approach performs better than rotation-based methods for translated datasets, where there is high correlation between utterance structure of training data in source language and evaluation data in target language. On the other hand, rotation-based alignments generalise better when the downstream target sentence distribution is dissimilar from the source sentence distribution, as is the case for non-translated NER tasks.

\subsection{Aligned Source Language vs./ Target Language Training}

\emph{FineTuneAlign${}_{src \to tgt}$} achieves top F1 score of 80.21 on SNIPS-Italian dataset which is not far from the score of 83 from a BERT-based model trained on 1400 manually-annotated Italian utterances~\shortcite{snips-it}. Also, our best alignment score of 80.90 for FB-Spanish (\emph{FineTuneAlign${}_{src \to tgt}$}) surpasses translate-train baseline~\shortcite{fb} where the annotations are automatically inferred from a NMT model. This suggests that for closer target languages, fine-tuning based alignment are not far behind from unaligned models trained on additional target language labelled examples.

Performance improvement from fine-tuning alignment for translated datasets should not be attributed to superficial transfer of entity information from source language. An evidence to support this claim is the strong performance on the SNIPS Italian-SF dataset, which has been translated from SNIPS dataset~\cite{snips-it}, where English entities have been replaced with Italian entities collected from the Web during dataset preparation. Therefore, during validation, the model came across utterances with similar structure but different entities, which shows that improvement from fine-tuning alignment is largely independent of language specific entity memorisation.

\section{Related Work}

\newcite{diab} propose to align ELMo embeddings ~\cite{peters} with word-level and sentence-level alignments. They compare the aligned ELMo with static character-level embeddings with similar alignments.

\newcite{cao} originally proposed fine-tuning alignment of mBERT language sub-spaces. They claim these methods are strictly stronger to rotation alignments methods based solely on zero-shot experimentation on XNLI task~\cite{xnli}, a semantic sentence-level classification task generated through translation from source language. On the contrary, we observe that fine-tuning does not improve performance across all tasks, particularly structural tasks, where utterance structure changes and there is higher incidence of domain shift. This raises the question whether translated datasets are biased to fine-tuning alignment, and whether such datasets are a good evaluation test-bed for general cross-lingual transfer.

\newcite{zwang} applies rotational alignment to mBERT and reports results on CoNLL NER tasks, however the main focus of their work is on the overlap of static bilingual embeddings. They do not extend similar analysis on contextualised embeddings. In our work, drawing from the observations made by \newcite{libovick2019languageneutral} on the distribution of languages in mBERT space, we propose a normalization mechanism to increase the overlap of two languages distributions prior to computing rotational alignment.

\newcite{schuster2019} originally proposed dictionary supervision to align ELMo with rotational transform. They claim supervision from dictionary is superior to using parallel word aligned corpora, however they do not substantiate these through comparative experiments. We observe that parallel corpus supervision is stronger than dictionary supervision possibly because of considering contextual alignment.

\section{Conclusion}
In this paper, we investigate cross-lingual alignment methods for multilingual BERT. We empirically evaluate their effect on zero-shot transfer for downstream tasks of two types: structural NER and semantic Slot-filling, across a set of diverse languages. Specifically, we compare rotation alignment and fine-tuning cross-lingual alignment. We compare the effect of dictionary and parallel corpora supervision across all tasks. We also propose a novel normalisation technique that improves state-of-the-art performance on zero-shot NER and Semantic Slot-filling downstream tasks, motivated by how languages are distributed across the mBERT space. Our experimental settings cover four datasets families (one for NER and three for SF) across eight language pairs. 

Key findings of this paper are as follows: (1) rotation-based alignments show large performance improvements (up to +19\% absolute for Turkish ATIS-SF) on moderately close languages, only a small improvement for very close target languages and no improvement for very distant languages; (2) we propose a novel normalisation which centers language distributions prior to learning rotation maps and is consistently shown to improve rotation alignment across all tasks particularly for Thai, by up to 3\% absolute; (3) rotational alignments are more robust and generalise well for structural tasks such as NER which may have higher utterance variability and domain shift; (4) supervision from parallel corpus generally leads to better alignment than dictionary-based, since it offers the possibility of generating contextualised alignments; (5) fine-tuning alignment improves performance for semantic tasks such as slot-filling where the source language data has minimal shift in utterance structure or domain from target language data and particularly improves performance for extremely distant languages (up to +8.39\% absolute higher for Thai FB-SF) compared to rotation alignment; (6) for close languages and tasks with similar utterance structure, zero-shot fine-tuning alignment is competitive versus unaligned models trained on additional annotated data in target language.

This work aims to pave the way for optimising language transfer capability in contextual multilingual models. In the future, we would like to further investigate patterns in the embedding space and apply alignment methods into specific regions of the multilingual hyperspace to obtain more tailor-suited alignments between language pairs. We would also like to evaluate zero-shot capabilities of alignments when applied to other language tasks.

\bibliographystyle{acl_natbib}
\bibliography{emnlp2020}

\end{document}